\pdfoutput=1
\documentclass[11pt]{article}

\usepackage[]{acl}

\usepackage{times}
\usepackage{latexsym}
\usepackage{multirow}
\usepackage{color,xcolor}
\usepackage{booktabs} 
\usepackage{arydshln}
\usepackage{amsmath}
\usepackage{enumitem}

\usepackage[T1]{fontenc}
\usepackage[utf8]{inputenc}

\usepackage{microtype}

\usepackage{inconsolata}
\usepackage{amsfonts}
\usepackage{graphicx}
\usepackage{subcaption}

\title{Legal Mathematical Reasoning with LLMs: Procedural Alignment through Two-Stage Reinforcement Learning}

\author{
    \bf Kepu Zhang\footnotemark[1]\textsuperscript{1},
    \bf Guofu Xie\footnotemark[1]\textsuperscript{1},
    \bf Weijie Yu\textsuperscript{2},\\
    \bf Mingyue Xu\textsuperscript{2},
    \bf Xu Tang\textsuperscript{2},
    \bf Yaxin Li\textsuperscript{2},
    \bf Jun Xu\textsuperscript{1} \\
    \textsuperscript{1}Gaoling School of Artificial Intelligence, Renmin University of China \\
    \textsuperscript{2} University of International Business and Economics \\
   \\
}

\begin{document}
\maketitle
{
\renewcommand{\thefootnote}{\fnsymbol{footnote}}
\footnotetext[1]{Equal Contribution.}
}
\begin{abstract}
Legal mathematical reasoning is essential for applying large language models (LLMs) in high-stakes legal contexts, where outputs must be both mathematically accurate and procedurally compliant. However, existing legal LLMs lack structured numerical reasoning, and open-domain models, though capable of calculations, often overlook mandatory legal steps.
To address this, we present \textbf{LexNum}, the first Chinese legal mathematical reasoning benchmark, covering three representative scenarios where each instance reflects legally grounded procedural flows.
We further propose \textbf{LexPam}, a two-stage reinforcement learning framework for efficient legal reasoning training. Leveraging curriculum learning, we use a stronger teacher model to partition data into basic and challenging subsets. A lightweight 1.5B student model is then fine-tuned with Group Relative Policy Optimization, which avoids costly value networks and enables stable training from sparse, end-of-sequence rewards. The first stage improves accuracy and format; the second introduces a novel reward to guide procedural alignment via task-specific legal elements.
Experiments show that existing models perform poorly on LexNum, while LexPam enhances both mathematical accuracy and legal coherence, and generalizes effectively across tasks and domains.
\end{abstract}
\section{Introduction}
Legal mathematical reasoning is a critical capability for applying large language models (LLMs) in real-world legal systems. This domain-specific task—spanning scenarios such as compensation calculation and liability apportionment—requires not only the precise application of formulas and parameters, but also adherence to legally prescribed procedures, such as confirming responsibility and determining applicable insurance. Failing either component—computation or procedure—can result in legally invalid or non-compliant outcomes. This dual requirement makes legal mathematical reasoning fundamentally more complex than general mathematical~\cite{cobbe2021training} or legal language tasks~\cite{yao2022leven}.

\begin{figure}[t]
    \centering
    \includegraphics[width=\linewidth]{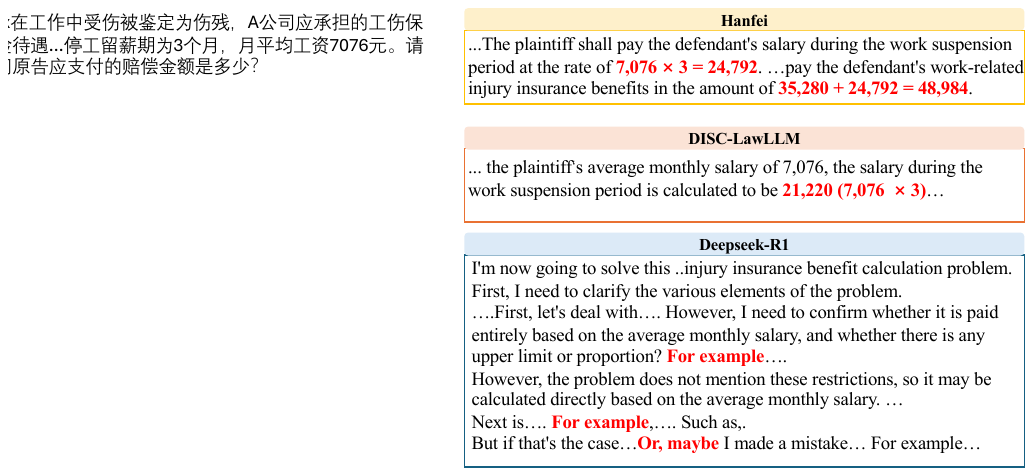}
    \caption{
    Responses to a work injury compensation case. Hanfei~\cite{HanFei} and DiscLaw-LLM~\cite{yue2023disclawllm} miscalculates; DeepSeek-R1~\cite{guo2025deepseek} is uncertain and lacks procedural alignment.
    }
    \label{fig:legal llm and r1}
    \vspace{-3mm}
\end{figure}

\begin{figure*}
    \centering
\includegraphics[width=0.98\linewidth]{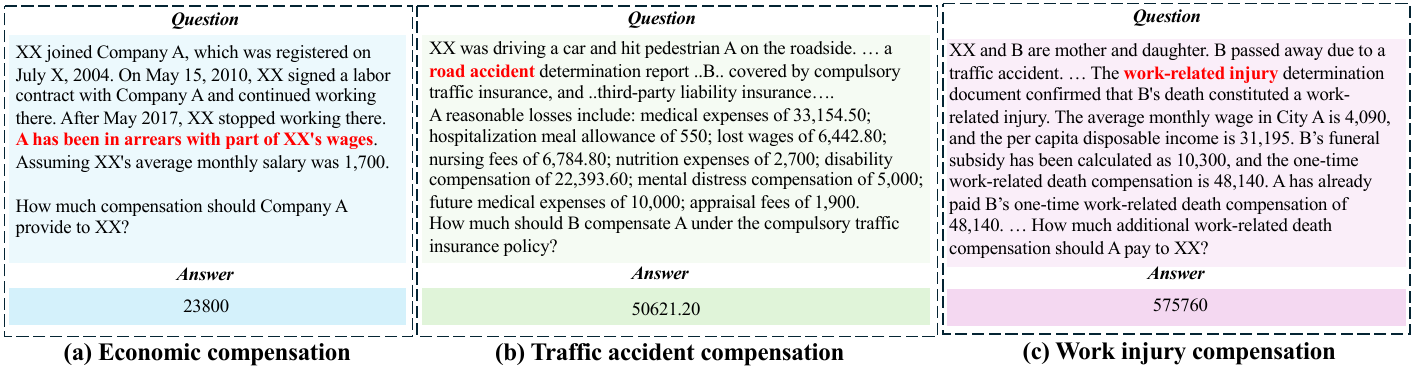}
    \caption{
    Examples from LexNum, our proposed benchmark for legal mathematical reasoning. Each scenario—(a) economic compensation, (b) traffic accident compensation, and (c) work injury compensation—requires multi-step reasoning that combines procedural logic with numerical calculation, reflecting real-world legal complexities.
   }
    
    \label{fig:example}
    \vspace{-3mm}
\end{figure*}

However, existing approaches fall short in meeting this requirement from two directions. Legal-domain LLMs, such as fuzi.mingcha~\cite{sdu_fuzi_mingcha} and DISC-LawLLM~\cite{yue2023disclawllm}, are fine-tuned for legal knowledge retention and textual generation, but lack the ability to perform accurate mathematical inference. As shown in Figure~\ref{fig:legal llm and r1}, these models may apply fixed formulas but miscalculate even simple operations, leading to numerically incorrect results. 
In contrast, open-domain reasoning models like DeepSeek-R1~\cite{guo2025deepseek} and OpenAI-o1~\cite{jaech2024openai} demonstrate strong step-by-step calculation abilities, yet fail to incorporate domain-specific legal logic. Their responses tend to over-explain or remain uncertain, often omitting essential statutory steps or legal constraints. As the bottom of Figure~\ref{fig:legal llm and r1} illustrates, DeepSeek-R1 may recognize the need for legal context but are unable to operationalize it, resulting in answers that are verbose, ambiguous, or procedurally invalid. These limitations underscore a critical gap: the need for models that can integrate both precise mathematical reasoning and legally grounded procedural logic.

Bridging this gap requires overcoming three key challenges.
\textbf{First}, 
Most prior benchmarks either focus on general legal question answering~\cite{zhong2020jec,chen2023equals} or abstract arithmetic reasoning~\cite{cobbe2021training,hendrycksmath2021}, making it difficult to assess how well a model integrates legal procedure with numerical computation. 
While recent datasets such as LexEval~\cite{li2024lexeval} and LawBench~\cite{fei2023lawbench} include criminal sentencing tasks, these typically involve single-step reasoning with discrete outputs (e.g., prison terms), which differ significantly from the multi-step, interdependent calculations required in civil disputes. For instance, traffic accident compensation involves liability assessment, insurance coverage, and layered monetary calculations. Without benchmarks reflecting such real-world complexity, model predictions are difficult to validate or deploy in practice.
\textbf{Second}, 
legal mathematical reasoning typically requires multi-step inference aligned with statutory procedures, yet fine-grained supervision over intermediate steps is rarely available. Annotating such reasoning paths
is not only costly and time-consuming, but also prone to inconsistency due to the ambiguity and variability in legal interpretation. This lack of step-level supervision poses a significant obstacle for model training, requiring methods that can learn effectively and stably from sparse, end-of-sequence reward signals without relying on explicitly annotated intermediate steps.
\textbf{Third}, 
aligning model behavior with legal reasoning requires training objectives that account for both numerical accuracy and procedural validity. However, most reward designs in existing reinforcement learning setups, such as the accuracy- and format-based rewards used in DeepSeek-R1, primarily target surface-level correctness and struggle to capture the structured logic mandated by legal norms. These reward functions provide insufficient guidance for producing responses that are not only correct in value, but also coherent with the legally required reasoning path as shown in Figure~\ref{fig:legal llm and r1}.

To address these challenges, we propose \textbf{LexNum}, the first Chinese benchmark explicitly designed for legal mathematical reasoning, and \textbf{LexPam}, a two-stage reinforcement learning framework for training LLMs under supervision constraints.
LexNum covers three high-impact civil law scenarios—economic compensation, work injury, and traffic accident cases—each requiring multi-step reasoning that combines legal procedural logic with arithmetic precision. All instances are constructed to reflect real-world procedural dependencies, enabling rigorous evaluation of a model’s ability to perform lawful and interpretable computation.

Building on this benchmark, LexPam introduces an efficient training strategy that aligns model outputs with legal reasoning goals while avoiding costly supervision. Inspired by curriculum learning, we use a stronger teacher model to partition the training data into basic and challenging subsets. A lightweight 1.5B student model is then fine-tuned using Group Relative Policy Optimization (GRPO)~\cite{guo2025deepseek} that eliminates the need for value functions and learns directly from sparse, end-of-sequence rewards. The first stage improves numerical accuracy and response formatting, while the second introduces a novel reward that encourages procedural alignment by promoting the inclusion of scenario-specific legal elements. This design enables training of LLMs for legally valid mathematical reasoning—without relying on manually annotated reasoning steps.
Extensive experiments on LexNum show that existing legal LLMs perform poorly, with average accuracy below 15\%. Our proposed LexPam achieves 60.65\% accuracy, surpassing DeepSeek-R1-7B (31.52\%) and GPT-4o-mini (56.93\%), and approaching the performance of much larger models such as QwQ-32B (70.94\%) and DeepSeek-R1-671B (73.42\%). These results highlight the effectiveness of our efficient and procedurally aligned training framework.

To summarize, our contributions are as follows:
\begin{itemize}[leftmargin=*]
\item We present LexNum, the first dataset explicitly designed for legal mathematical reasoning in Chinese, spanning three representative civil law scenarios: economic compensation, work injury, and traffic accident cases.
\item We introduce LexPam, a reinforcement learning framework that incorporates legal procedural awareness into reward design, enabling LLMs to generate answers that are both mathematically accurate and procedurally compliant.
\item We conduct comprehensive evaluations across legal- and reasoning-specific LLMs. Results show that existing models perform poorly on LexNum, while LexPam significantly improves legal mathematical reasoning, even outperforming models several times larger in scale.
\end{itemize}

\section{Related Work}
\subsection{Legal LLM and Benchmark Dataset}
Legal LLMs are typically adapted from open-domain models through domain-specific pretraining or fine-tuning using legal statutes, case documents, synthetic legal QA pairs, and annotated task-specific datasets~\cite{yue2023disclawllm,wisdomInterrogatory,LAWGPT-zh,HanFei}. These adaptations improve factual grounding and enable more relevant responses in legal applications.

Several benchmarks evaluate legal LLMs from different perspectives, including LexEval~\cite{li2024lexeval}, LawBench~\cite{fei2023lawbench}, and CitaLaw~\cite{zhang2024citalaw}, which assess capabilities across judgment prediction, legal comprehension, citation extraction, and general legal QA. Recent studies have also begun to explore legal reasoning~\cite{yu2025evaluating,deng2024enabling,zhang2024beyond}, but most focus on classification-oriented tasks such as judgment prediction or reading comprehension, rather than structured numerical reasoning.

While some prior benchmarks include legally motivated calculations—e.g., the “Penalty Prediction” task in LexEval and “Prison Term Prediction” tasks in LawBench, derived from CAIL-2018~\cite{xiao2018cail2018}—they are limited in scope and reasoning depth. These tasks typically involve discrete outputs and can often be solved via single-step application of criminal statutes.

In contrast, our proposed dataset LexNum differs in three key ways:
\begin{itemize}[leftmargin=*]
\item \textbf{Domain scope}: LexNum targets civil law scenarios, including economic compensation, workplace injury, and traffic accident cases—practical domains central to everyday legal consultation.
\item \textbf{Reasoning depth}: Unlike single-step prison term prediction, LexNum requires multi-step computation guided by statutory rules, such as liability ratio determination and conditional insurance application.
\item \textbf{Answer format}: LexNum involves continuous-valued monetary outcomes with arithmetic operations (e.g., multiplication, division, percentage), moving beyond discrete classification.
\end{itemize}
By addressing legal mathematical reasoning in high-frequency civil contexts, LexNum fills a critical gap in current benchmarks and provides a foundation for developing legal LLMs that can support more realistic, procedurally grounded legal service.

\subsection{Reasoning}
The emergence of reasoning-focused LLMs, such as DeepSeek-R1~\cite{guo2025deepseek} and OpenAI-o1~\cite{jaech2024openai}, has led to rapid progress in open-domain reasoning. Recent work has explored reward shaping to constrain unnecessarily long reasoning chains~\cite{aggarwal2025l1,luo2025o1}, efficient token-level reasoning compression~\cite{han2024token}, and boosting mathematical performance using limited high-quality data~\cite{muennighoff2025s1,ye2025limo}. Other efforts have focused on knowledge distillation~\cite{li2023mixed,zhu2024improving}, transferring reasoning capabilities from large to smaller models for efficiency.
However, this body of work focuses almost exclusively on open-domain tasks, such as math problems or code-based reasoning. In contrast, legal mathematical reasoning introduces domain-specific procedural constraints that open-domain models fail to capture. To our knowledge, this work is the first to explore LLM reasoning under legal constraints, bridging open-domain reasoning and real-world legal applications.

\section{The Proposed Dataset: LexNum }\label{sec:dataset}
To construct LexNum, we collected real-world legal documents from \texttt{pkulaw}\footnote{https://www.pkulaw.com}, covering three high-frequency litigation domains in Chinese civil law: economic compensation, workplace injury, and traffic accident cases. Each document includes structured metadata—case background, legal process, and final compensation result—and is pre-anonymized. We developed a two-stage pipeline involving LLM-assisted extraction and selective human verification to efficiently construct high-quality legal mathematical reasoning examples. Details of each subtask are provided in Appendix~\ref{sec:appendix task}.

\subsection{LLM-Based Content Extraction}
Raw legal documents contain substantial information irrelevant to mathematical reasoning. Rather than rely on fully manual annotation—accurate but prohibitively expensive—we adopt a lightweight LLM-based extraction approach. While deep reasoning models could potentially identify scattered reasoning chains, they are computationally inefficient. Instead, we use GPT-4o, which provides sufficient comprehension without heavy inference cost.
Given a case document and relevant statutory provisions, GPT-4o is prompted to extract structured question-answer pairs related to legal mathematical reasoning. This includes identifying the numerical query, applicable rules, and the final outcome. Prompting strategies and examples are included in Appendix~\ref{sec:appendix prompt}.

\subsection{Efficient Quality Assurance}\label{sec:data quality}

To ensure quality while minimizing human labor, we adopt an LLM-then-human filtering strategy. First, GPT-4o is used to screen generated examples for completeness and internal consistency—i.e., whether the provided information suffices to derive the answer. Examples flagged as incomplete are passed to human reviewers. In addition, a small proportion of LLM-approved examples are also sampled for verification, as LLM outputs may contain subtle legal or numerical errors.

We employ three reviewers with legal training for manual review. Initially, each annotator reviewed one full dataset, followed by a cross-checking phase where each reviewed 50\% of the other two. A final audit was conducted by randomly sampling 50 examples per dataset. If any errors were found during the audit, a full re-check of the corresponding dataset was triggered. During annotation, reviewers corrected mismatches and refined unclear reasoning steps when needed. Full details are in Appendix~\ref{sec:appendix annotator}.

\section{The Proposed Method: LexPam}
In this section, we present LexPam, a two-stage reinforcement learning framework comprising curriculum-based data selection (§\ref{sec:selection}), foundational training for legal mathematical reasoning (§\ref{sec:rl-1}), and procedural alignment through targeted reward optimization (§\ref{sec:rl-2}).

\subsection{Curriculum-Based Data Selection}
\label{sec:selection}
To optimize training efficiency under limited computational resources, we focus on fine-tuning a 1.5B reasoning LLM. Since the model exhibits varying performance across samples of different difficulty, we adopt a curriculum-based data partitioning strategy guided by a stronger 7B LLM (DeepSeek-R1-Distill-Qwen-7B). Specifically, we use the 7B model to perform inference over the training set. Samples it answers correctly are assigned to the first-stage training set $\mathcal{D}_1$, which emphasizes basic numerical computation and output format standardization. The remaining samples, which the 7B model fails to solve, form the second-stage training set $\mathcal{D}_2$, aimed at improving the model’s ability for procedural alignment. This design provides a difficulty-aligned learning trajectory for the 1.5B model.

\textbf{Discussion:} We choose a 7B model rather than a larger or smaller alternative for two reasons: (1) Larger models introduce a significant performance gap, making their solvable samples too difficult for the 1.5B model to learn from effectively. (2) Smaller models, including the target 1.5B itself, provide insufficient guidance and cannot meaningfully differentiate between easy and hard samples for curriculum construction.

\subsection{Stage-1: Foundation Training}
\label{sec:rl-1}
The first stage focuses on low-complexity legal mathematical tasks to build the model’s capability in structured computation and standardized legal output formatting. We train on $\mathcal{D}_1$, the subset of samples solved by the teacher model, which serve as reliable examples for foundational learning.

Each input consists of a user query describing a compensation calculation, along with relevant facts. The model generates reasoning steps and a final result, where the intermediate reasoning chain is enclosed in ``\texttt{<think></think>}'' and the final answer in \texttt{boxed\{\}}. Training is performed using the GRPO algorithm~\cite{shao2024deepseekmath}. We define the corresponding reward as:
\begin{equation}
\label{eq:reward1}
r_1 = r_\mathrm{correct} + \alpha \cdot r_\mathrm{format},
\end{equation}
where $r_\mathrm{correct}$ indicates whether the final boxed result matches the reference answer, and $r_\mathrm{format}$ evaluates whether the output adheres to the required structure. The hyperparameter $\alpha$ balances the focus between correctness and formatting.

\subsection{Stage-2: Procedural Grounding}
\label{sec:rl-2}
In this stage, we aim to instill procedural legal reasoning by training on $\mathcal{D}_2$, the subset of examples unsolved by the teacher. These samples typically require deeper legal understanding, domain-specific terminology, and adherence to statutory steps.

Building on the first stage’s foundation of numerical correctness and structured output, we introduce an additional reward component, $r_\mathrm{law}$, to promote the use of legally appropriate terminology and reasoning patterns.
To operationalize this, $r_\mathrm{law}$ measures the presence of key legal elements in model outputs. These elements vary by task type, based on real-world legal procedures:
\begin{itemize}[leftmargin=*]
\item Economic: \texttt{Compensation Type}, \texttt{Monthly Calculation}, \texttt{Compensation Calculation}
\item Work Injury: \texttt{Injury Recognition}, \texttt{Liability}, \texttt{Benefit Calculation}, \texttt{Insurance}, \texttt{Compensation Calculation}
\item Traffic Accident: \texttt{Liability}, \texttt{Insurance}, \texttt{Compensation Calculation}
\end{itemize}
To quantify this, we define a task-specific set with size $N_j$ of legal keywords $\{k_{j,1}, \ldots, k_{j,N_j}\}$ for each task type $j$ (e.g., economic compensation, work injury, or traffic accident). The procedural reward $r_\mathrm{law}$ is computed as the weighted coverage of these keywords in the model’s output $R$:
\begin{equation}
\label{eq:reward law}
r_{law}=\sum_{i=1}^{N_j}\omega _{j,i}\cdot \mathbb{I}(k_{j,i},R) ,
\end{equation}
where $\omega_{j,i} \in [0,1]$ is the importance weight assigned to the $i$-th keyword of task $j$, and $\mathbb{I}(\cdot)$ is the indicator function that evaluates to 1 if the keyword appears in $R$, and 0 otherwise. For simplicity, we use uniform weighting in our experiments. That is, we set $\omega _{j,i}=\frac{1}{N_j}$.

When the three datasets are merged into a unified training corpus (see §\ref{sec:diversity}), we first identify the target scenario from the input and compute $r_\mathrm{law}$ using the corresponding keyword set.

The full reward function for this stage combines correctness, formatting, and procedural alignment:
\begin{equation}
\label{eq:reward2}
r_2 = r_\mathrm{correct} + \alpha \cdot r_\mathrm{format} + \beta \cdot r_\mathrm{law},
\end{equation}
where $\beta$ is the strength of procedural alignment.

\textbf{Discussion:} Because annotating high-quality legal reasoning paths is expensive and error-prone, we do not apply supervised fine-tuning (SFT) before RL. Instead, we directly train a distilled model using GRPO on the above reward function, improving efficiency and avoiding dependence on step-level annotations.

\section{Experiments}\label{sec:experiment}
\subsection{Experimental Settings}
\subsubsection{Datasets and Metrics}
We evaluate models on our proposed legal mathematical reasoning benchmark, LexNum, which consists of three task-specific subsets: Economic Compensation (EC), Work Injury Compensation (WC), and Traffic Accident Compensation (TC). Following standard practice in open-domain mathematical reasoning~\cite{cobbe2021training,muennighoff2025s1}, we report accuracy as the primary metric, measuring whether the final computed result exactly matches the ground truth. Dataset statistics are summarized in Table~\ref{tab:data_main}.

\begin{table}[t]
\centering
\resizebox{\columnwidth}{!}{
\begin{tabular}{lccccc}
    \toprule
    \multirow{1}{*}{\textbf{Dataset}} &\multirow{1}{*}{\#\textbf{Train}} &\multirow{1}{*}{\#\textbf{Test}} &\multirow{1}{*}{\textbf{Avg\_Train\_Len}} 
    &\multirow{1}{*}{\textbf{Avg\_Test\_Len}}\\
    \midrule
    \textbf{EC} &1796&450&184.65&183.63 \\
    \textbf{WC} &774&194&168.79&170.16\\
    \textbf{TC}&395&99&194.29&192.59 \\
    \bottomrule
\end{tabular}}
\caption{Statistics of our LexNum. \textbf{\#Train}, \textbf{\#Test} denote the number of the train and test datasets. \textbf{Avg\_Train\_Len} and \textbf{Avg\_Test\_Sent} represent the average length of the train and the test datasets queries.}
\label{tab:data_main}
% \vspace{-3mm}
\end{table}

\subsubsection{Evaluation Models}
We selected both legal domain-specific LLMs and reasoning LLMs to investigate their performance on legal mathematical reasoning tasks.  

For legal LLMs we chose seven models: \textbf{Zhihai}~\cite{wisdomInterrogatory}, \textbf{fuzi.mingcha}~\cite{sdu_fuzi_mingcha}, \textbf{DISC-LawLLM}~\cite{yue2023disclawllm}, \textbf{LawGPT\_zh}~\cite{LAWGPT-zh}, \textbf{Tailing}\footnote{https://github.com/DUTIR-LegalIntelligence/Tailing}, \textbf{LexiLaw}\footnote{https://github.com/CSHaitao/LexiLaw}, and \textbf{HanFei}\cite{HanFei}. These models have been extensively trained on legal datasets and possess substantial knowledge of legal scenarios.  

We fine-tuned our models based on DeepSeek-R1-Distill-Qwen-1.5B~\cite{guo2025deepseek} and compared them with \textbf{GRPO-Base}~\cite{shao2024deepseekmath}, which performs direct GRPO training using rewards that consider only response format and calculation results. Our proposed methods include \textbf{GRPO-Law}, which directly incorporates rewards based on legal procedures, and \textbf{LexPam}, which uses a two-stage RL training approach.
For larger-scale LLMs, we evaluated models distilled by DeepSeek, specifically DeepSeek-R1-Distill-Qwen-7B (\textbf{DeepSeek-R1-7B}), DeepSeek-R1-Distill-Qwen-32B (\textbf{DeepSeek-R1-32B}), and \textbf{DeepSeek-R1} itself.
In addition, we included \textbf{QwQ-32B-Preview}~\cite{qwq-32b-preview} and \textbf{GPT-4o-mini} in our evaluation. Appendix~\ref{sec:appendix llm} provides more details (URLs and licenses).

\subsubsection{Implementation Details}
We conducted training and testing on dual A6000 GPUs.  
For GRPO training, we set the learning rate to 1e-6, the number of generations (num\_generations) to 4, and the maximum completion length to 768. We used LoRA for efficient fine-tuning of the LLM, with LoRA parameters set to r is 16 and alpha is 16. The hyperparameters $\alpha$ and $\beta$ were both set to 0.1.  
For testing, we set the temperature to 0 to ensure reproducibility.  
We trained the model using DeepSpeed ZeRO-2~\cite{rajbhandari2020zero} and accelerated inference with vLLM~\cite{kwon2023efficient}.

\subsection{Main Results}

\begin{table}[t]
\centering
    % \vspace{-3px}
    \resizebox{0.48\textwidth}{!}{
         \begin{tabular}{cc|ccc|c}
          \toprule
          Model & Scale & EC & WC  & TC  &\textbf{Avg}\\
          % \midrule
          \hline
            \multicolumn{6}{c}{\textbf{Legal LLM}}\\
            \hline
          % &\multicolumn{6}{c}{\shortstack{Legal LLM}}\\
DISC-LawLLM &13B & 12.44 & 2.58 & 18.18 & 11.07 \\
fuzi.mingcha &6B & 14.00 & 4.12 & 10.10 & 9.41 \\
LexiLaw &6B  & 4.67 & 2.06 & 5.05 & 3.93\\
Tailing &7B & 18.89 & 10.31 & 15.15 & 14.78\\
zhihai &7B & 6.00 & 3.09 & 5.05 & 4.71 \\
LawGPT\_zh &6B & 5.11 & 3.61 & 5.05 & 4.59\\
HanFei &7B & 7.56 & 1.55 & 8.08 & 5.73 \\
\hline
\hline
\multicolumn{6}{c}{\textbf{Small-Scale LLM (<10B)}}\\
\hline
Deepseek-R1-1.5B &1.5B & 16.00 & 22.16 & 18.18 & 18.78 \\
Deepseek-R1-7B &7B & 32.22 & 36.08 & 26.26 & 31.52 \\
GRPO-Base&1.5B & 52.00 & 61.86 & 38.38 & 50.75 \\
\cdashline{1-6}
GRPO-Law&1.5B & 59.11 & 66.49 & 46.46 & 57.36 \\
LexPam&1.5B & \textbf{64.89} & \textbf{69.59} & \textbf{47.47} & \textbf{60.65} \\
\hline
\hline
\multicolumn{6}{c}{\textbf{Larger-Scale LLM ($\geq$10B)}}\\
\hline
Deepseek-R1-32B &32B & 70.22 & 68.56 & 48.48 & 62.42 \\
GPT-4o-mini &- & 60.00 & 59.28 & 51.52 & 56.93 \\
QwQ-32B-Preview &32B & 73.33 & 79.90 & 59.60 & 70.94 \\
Deepseek-R1 &671B & 74.67 & 81.96 & 63.64 & 73.42 \\
          \bottomrule
        \end{tabular}
}
        \caption{Performance Comparison. LexPam (1.5B) aoutperforms all legal LLMs and small-scale reasoning models, and rivals larger-scale models. Bold indicates the best score within the <10B and the legal LLM groups.}
    % \vspace{-4mm}
    \label{tab:main results}
\end{table}

Table~\ref{tab:main results} presents accuracy scores on the LexNum benchmark, categorized into (1) legal-domain LLMs, (2) reasoning LLMs below 10B, and (3) larger-scale reasoning LLMs. We highlight the following findings:

\paragraph{LexPam outperforms all legal LLMs and small-scale reasoning models.}
LexPam achieves an average accuracy of 60.65\%, substantially outperforming all legal-domain models as well as all reasoning models $\leq$10B, such as DeepSeek-R1-1.5B (18.78\%). 
This demonstrates the effectiveness of LexPam’s two-stage training strategy in teaching legal procedural reasoning even under tight model size constraints.

\paragraph{LexPam rivals or exceeds larger-scale models.}
Despite being 1.5B in size, LexPam surpasses GPT-4o-mini (56.93\%) and comes close to 32B-scale models like DeepSeek-R1-32B (62.42\%) and QwQ-32B (70.94\%). This confirms the value of our curriculum-based data selection and procedural-aware reward design in improving sample efficiency and generalization.

\paragraph{Legal LLMs struggle with numerical reasoning.}
Most legal-domain LLMs perform poorly, often below 15\% accuracy. These models are typically fine-tuned on legal corpora for knowledge retention but lack explicit training for multi-step numerical inference. As shown in Figure~\ref{fig:legal llm and r1}, even basic arithmetic steps are frequently incorrect, suggesting that legal pretraining alone is insufficient for reasoning-intensive tasks.

\subsection{Ablation Study}
\begin{table}[t]
\centering
    % \vspace{-3px}
    \resizebox{0.43\textwidth}{!}{
         \begin{tabular}{c|ccc|c}
          \toprule
           Model & EC & WC  & TC  &\textbf{Avg}\\
\hline
GRPO-Base & 52.00 & 61.86 & 38.38 & 50.75 \\
GRPO-Law & 59.11 & 66.49 & 46.46 & 57.36 \\
\cdashline{1-5}
$\mathcal{D}_1$-Only & 56.00 & 57.22 & 37.37 & 50.20 \\
$\mathcal{D}_2$-Only & 57.33 & 65.46 & 44.44 & 55.75 \\
\cdashline{1-5}
LexPam & \textbf{64.89} & \textbf{69.59} & \textbf{47.47} & \textbf{60.65} \\
          \bottomrule
        \end{tabular}
        }
        \caption{Ablation results. LexPam outperforms all variants, demonstrating the effectiveness of curriculum staging and procedural rewards.}
    % \vspace{-4mm}
    \label{tab:aba results}
\end{table}

To assess the contribution of each design component in LexPam, we conduct an ablation study using the DeepSeek-R1-Distill-Qwen-1.5B model and the GRPO training framework. All models are evaluated on the full LexNum benchmark.

We compare the following variants:
\begin{itemize}[leftmargin=*]
\item \textbf{GRPO-Base}: Trained on $\mathcal{D}_1 \cup \mathcal{D}_2$ using only the base reward from Eq.\ref{eq:reward1} (correctness + format).
\item \textbf{GRPO-Law}: Same data as GRPO-Base, but uses the full reward from Eq.\ref{eq:reward2}, including legal procedural alignment.
\item \textbf{$\mathcal{D}_1$-Only} and \textbf{$\mathcal{D}_2$-Only}: Trained on $\mathcal{D}_1$ and $\mathcal{D}_2$ respectively , using the base reward.
\end{itemize}

From Table\ref{tab:aba results}, we observe that LexPam outperforms all variants, achieving 60.65\% average accuracy. Compared to GRPO-Base (50.75\%), LexPam gains +9.9 points, confirming the value of both curriculum structuring and legal-aware reward design. GRPO-Law (+6.6 over GRPO-Base) shows that procedural rewards alone provide a strong signal even without curriculum scheduling.

Interestingly, $\mathcal{D}_2$-Only outperforms $\mathcal{D}_1$-Only by a wide margin (+5.5), aligning with findings from prior work~\cite{ye2025limo} that training on harder samples promotes more transferable reasoning skills. However, LexPam still surpasses $\mathcal{D}_2$-Only by +4.9, demonstrating that gradual learning from easier samples (via $\mathcal{D}_1$) further stabilizes and strengthens model performance.

These results validate LexPam’s two-stage RL design: combining curriculum progression with procedural reward shaping yields consistently stronger legal reasoning.

\subsection{Cross-Domain Generalization}
To evaluate the generalization ability of our method beyond in-domain scenarios, we assess whether models trained on one legal domain can transfer effectively to others. Specifically, we conduct cross-domain experiments by training on one of the three LexNum subsets—Economic Compensation (EC), Work Injury Compensation (WC), or Traffic Accident Compensation (TC)—and testing on the remaining tasks.

\begin{figure}[t]
    \centering

    % \vspace{0.5em}  
    \begin{subfigure}{0.88\linewidth}
        \centering
        \includegraphics[width=\linewidth]{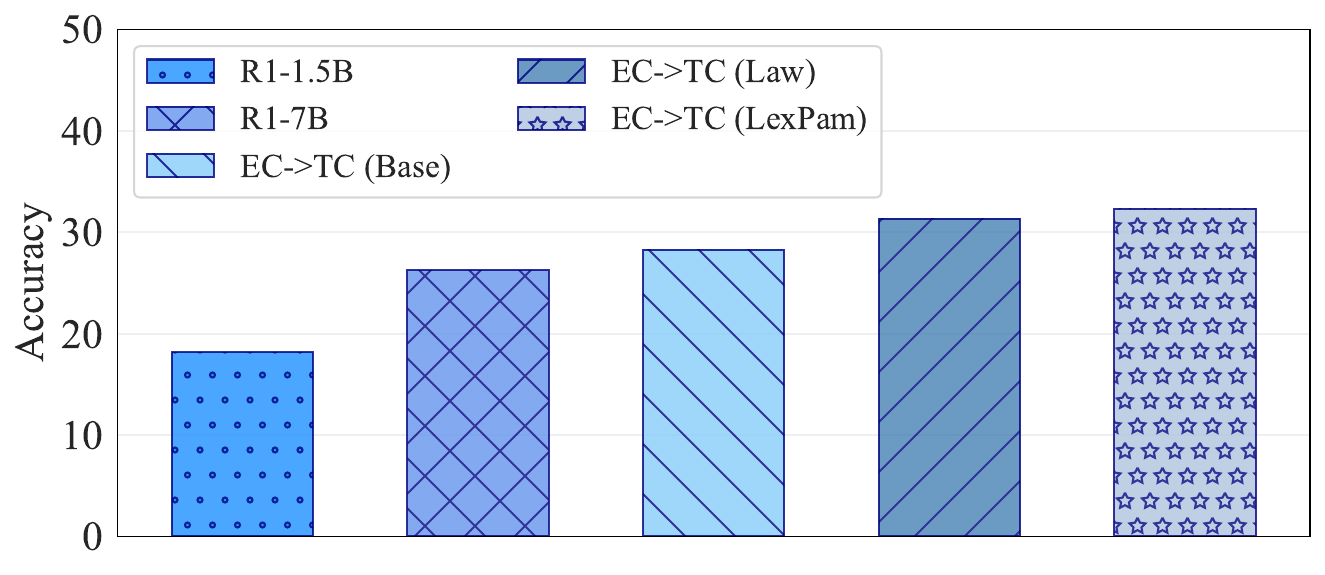}
        \subcaption{Train on EC, test on TC.}
    \end{subfigure}
    \hfill
    \begin{subfigure}{0.88\linewidth}
        \centering
        \includegraphics[width=\linewidth]{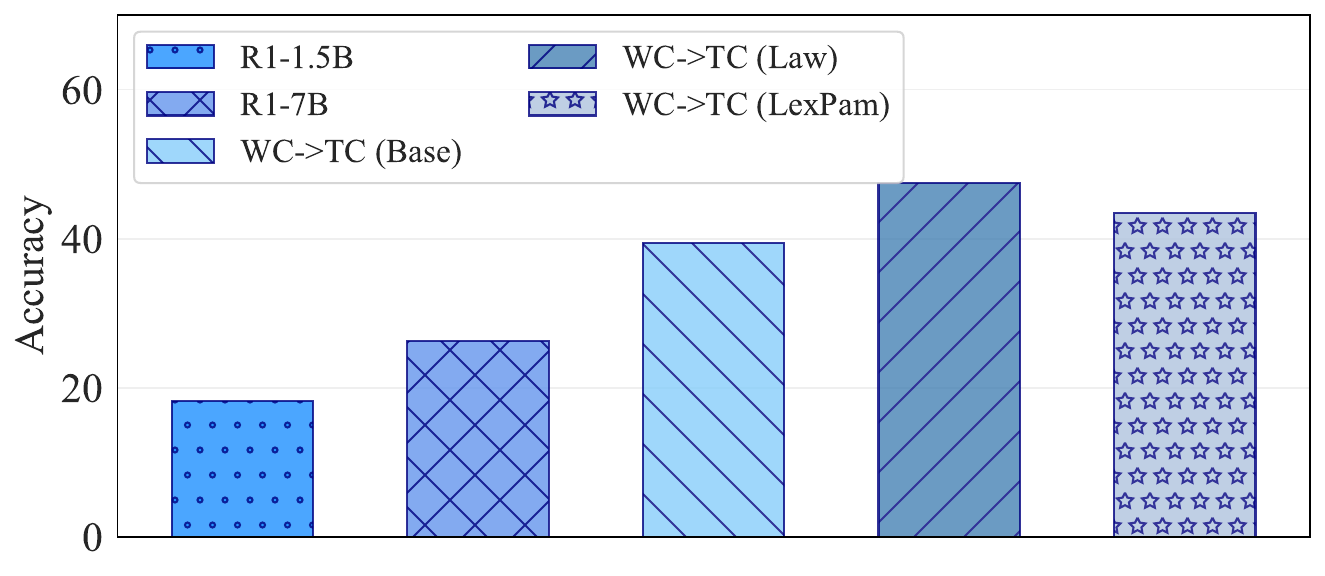}
        \subcaption{Train on WC, test on TC.}
    \end{subfigure}
        \begin{subfigure}{0.88\linewidth}
        \centering
        \includegraphics[width=\linewidth]{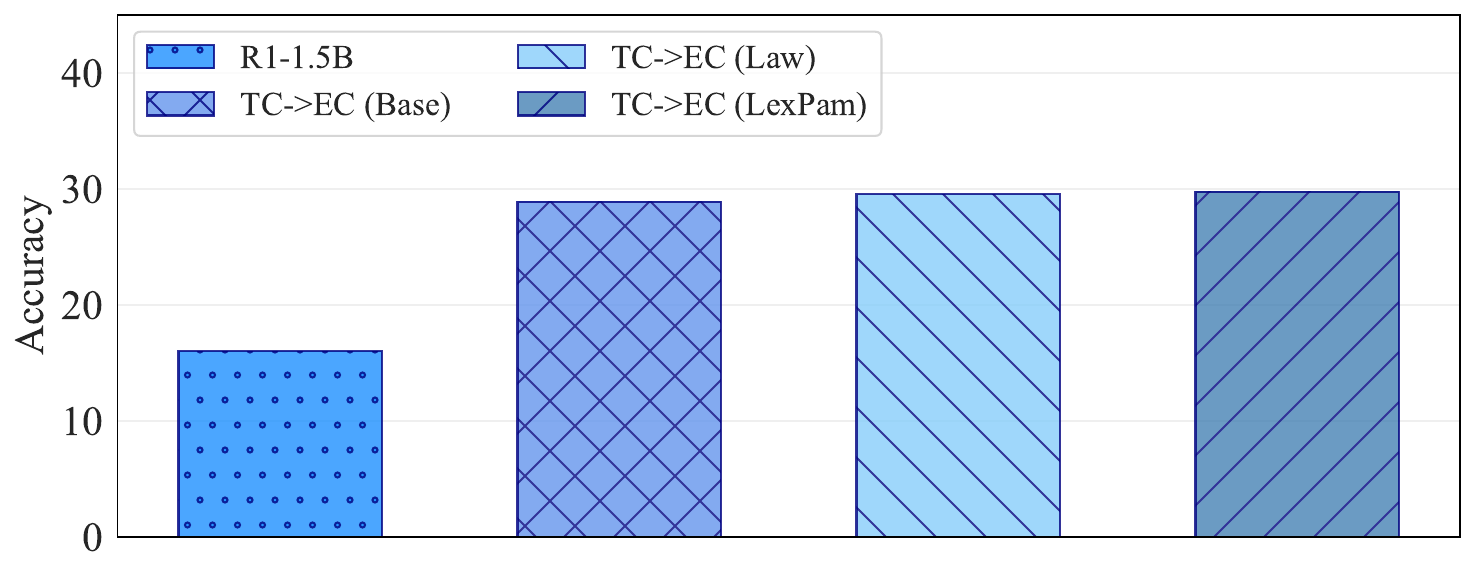}
        \subcaption{Train TC, test on EC}
    \end{subfigure}
    \hfill
    \begin{subfigure}{0.88\linewidth}
        \centering
        \includegraphics[width=\linewidth]{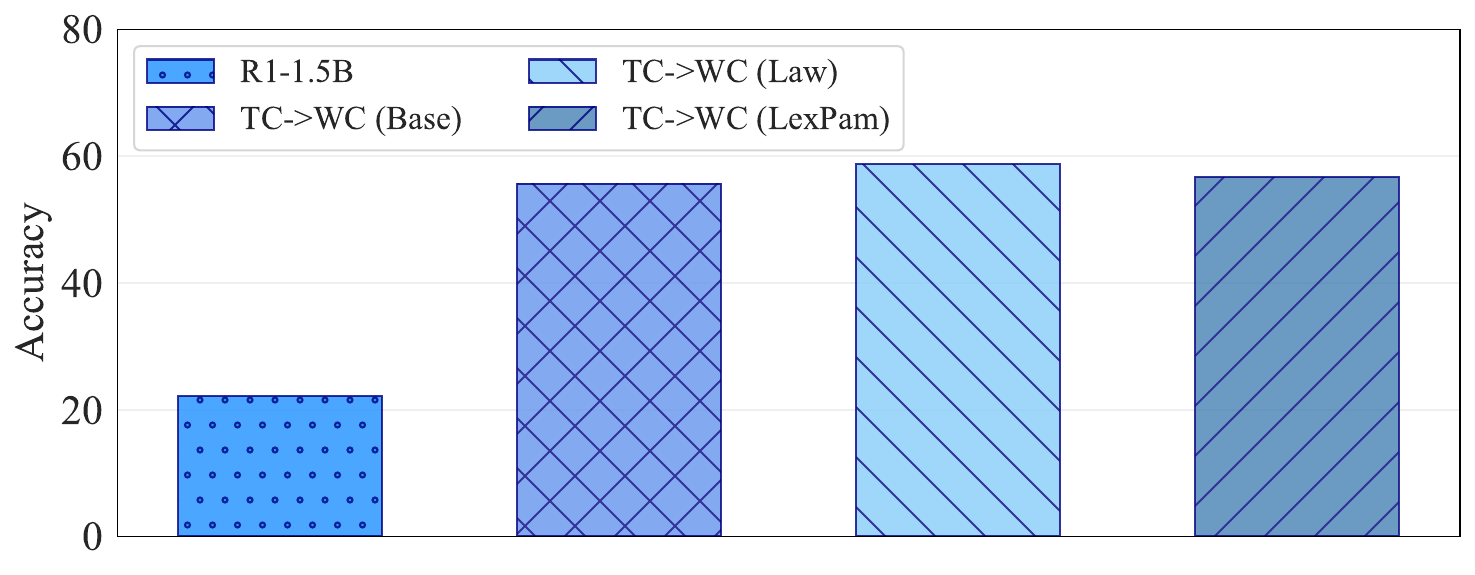}
        \subcaption{Train on TC, test on WC}
    \end{subfigure}
    \caption{
        Cross-domain Results. TC, EC, WC are short for traffic compensation, economic compensation, work injury compensation.
        R1-1.5B (7B) corresponds to DeepSeek-R1-Distill-Qwen-1.5B (7B). 
         LexPam shows strong generalization, matching or surpassing larger models.
    }
    \label{fig:cross}
    % \vspace{-3mm}
\end{figure}

As shown in Figure~\ref{fig:cross}, LexPam consistently outperforms GRPO-Base and performs on par with or better than larger-scale models such as DeepSeek-R1-7B, despite having only 1.5B parameters. This pattern holds across all training–testing configurations (EC$\rightarrow$TC, WC$\rightarrow$TC, TC$\rightarrow$EC, TC$\rightarrow$WC), demonstrating that the procedural-aware training strategy of LexPam enables strong cross-task transfer.

Notably, GRPO-Law also improves over GRPO-Base in most cases, further validating the effectiveness of the legal reward design. The ability of LexPam and GRPO-Law to maintain high accuracy across varying domains—despite differences in data distributions and legal formulations—highlights their robustness.

\subsection{Generalization via Task Merging}\label{sec:diversity}
Although the three LexNum tasks differ in legal provisions and calculation logic, they all center on compensation reasoning. We investigate whether merging these tasks can yield synergistic effects and enhance model generalization.
\begin{table}[t]
\centering
    % \vspace{-3px}
    \resizebox{0.45\textwidth}{!}{
         \begin{tabular}{c|ccc|c}
          \toprule
           Model & EC & WC  & TC  &\textbf{Avg}\\
          % \midrule
\hline
LexPam & 64.89 & 69.59 & 47.47 & 60.65 \\
\cdashline{1-5}
All-GRPO & 64.67 & 68.04 & 47.47 & 60.06 \\
All-GRPO-$\mathcal{D}_1$ & 60.22 & 59.28 & 42.42 & 53.97 \\
All-GRPO-$\mathcal{D}_2$ & 61.78 & 69.59 & 47.47 & 59.61 \\
\cdashline{1-5}
All-LexPam & \textbf{70.89} & \textbf{71.13} & \textbf{52.53} & \textbf{64.85} \\

          \bottomrule
        \end{tabular}
    }
        \caption{Results of task-merging experiments. All-LexPam achieves the best generalization across domains, demonstrating its scalability.
        }
    % \vspace{-3mm}
    \label{tab:diver results}
\end{table}
We evaluate the following multi-task training strategies:
\begin{itemize}[leftmargin=*]
\item \textbf{All-GRPO}: GRPO with base reward, trained on the full merged dataset.
\item \textbf{All-GRPO-$\mathcal{D}_1$ / $\mathcal{D}_2$}: Trained only on merged easy or hard samples, respectively.
\item \textbf{All-LexPam}: LexPam applied to merged data—trained on $\mathcal{D}_1$ with base reward, then on $\mathcal{D}2$ with scenario-specific legal rewards ($r\text{law}$ determined by task type).
\end{itemize}

As shown in Table~\ref{tab:diver results}, All-GRPO already surpasses the original LexPam, suggesting that training on diverse but structurally related legal tasks provides transferable inductive signals. All-LexPam achieves the best overall performance (64.85\%), confirming that LexPam’s two-stage structure scales effectively under task merging.

These results highlight the generalization potential of LexPam: its curriculum and procedural alignment mechanisms not only improve in-domain reasoning but also extend to heterogeneous legal settings through unified training.

\subsection{Human Evaluation}
To assess the procedural quality and user-facing suitability of model outputs, we conduct a human evaluation on responses generated for legal mathematical reasoning.
\begin{table}[t]
\centering
    % \vspace{-3px}
    \resizebox{0.45\textwidth}{!}{
         \begin{tabular}{c|ccc|c}
          \toprule
           Model & Completeness & Coverage  & Conciseness  &\textbf{AVG} \\
           % & Acc\\
          % \midrule
\hline
DISC-LawLLM & 3.89 & 3.92 & 4.58 & 4.13 \\
Tailing & 3.19 & 2.94 & \textbf{4.74} & 3.62  \\
Deepseek-R1 & 4.62 & 4.75 & 3.63 & 4.33 \\
\cdashline{1-5}
LexPam & \textbf{4.69} & \textbf{4.76} & 4.66 & \textbf{4.70}  \\
          \bottomrule
        \end{tabular}
    }
        \caption{Human evaluation results.}
    % \vspace{-3mm}
    \label{tab:case study}
\end{table}
We randomly sampled 40 queries from the traffic accident compensation dataset. For each query, we collected responses from four models: DISC-LawLLM, Tailing, DeepSeek-R1, and LexPam. The responses were anonymized and randomly ordered. Each query and its four associated responses were presented together to four independent annotators—Chinese law students with domain knowledge (distinct from those involved in §\ref{sec:data quality}).

Annotators rated each response on a 5-point Likert scale (1–5, higher is better) along three criteria:
\begin{itemize}[leftmargin=*]
\item \textbf{Completeness}: Are all key steps—liability, insurance, and compensation—present?
\item \textbf{Coverage}: Are the relevant legal elements from the question appropriately addressed?
\item \textbf{Conciseness}: Is the response clear and easy to read without unnecessary verbosity?
\end{itemize}

Table~\ref{tab:case study} shows the average scores across all raters. LexPam achieves the highest overall rating (4.70), with strong performance across all dimensions. Compared to DeepSeek-R1, which tends to over-explain, LexPam provides legally grounded yet concise outputs. These results suggest that LexPam not only improves procedural fidelity but also enhances user-oriented response quality.

Additional examples and evaluation details are available in Appendix~\ref{sec:appendix case study}.

\section{Conclution}
We present LexNum, the first benchmark for legal mathematical reasoning in Chinese, and introduce LexPam, a two-stage reinforcement learning framework that improves both computational accuracy and procedural compliance. LexPam leverages curriculum-based data selection and a novel legal-aware reward, enabling a lightweight 1.5B model to outperform a range of legal-domain and reasoning-focused LLMs.
Extensive experiments demonstrate LexPam’s effectiveness across in-domain and cross-domain settings, as well as in multi-task training scenarios. Human evaluation further confirms its strength in producing procedurally sound and user-readable outputs. 

\section{Limitations}
In this paper, due to computational resource limitations, we primarily conduct experiments on a 1.5B-parameter reasoning model to explore how to enhance its legal mathematical reasoning capabilities. In future work, when more computational resources become available, we plan to test LexPam on larger-scale LLMs to investigate whether its effectiveness holds.
Moreover, we have not yet annotated the legal mathematical reasoning process with Chain-of-Thought (CoT) in this paper, as producing high-quality CoT annotations for legal reasoning involves significant human effort. Due to the domain-specific nature of the task, such annotations require careful work by legal professionals. In the future, we aim to enrich the LexNum dataset by annotating the CoT reasoning processes that lead to the final answers, thereby further advancing research in the field of legal AI.

\section{Ethical Considerations}
The LexNum dataset we provide is intended to facilitate research on reasoning LLMs in the legal NLP domain. The data was collected from publicly accessible websites\footnote{https://www.pkulaw.com}, and all cases have been anonymized. During the dataset construction process, we also engaged legal professionals to review the dataset to ensure its compliance with legal and ethical standards.

\bibliography{custom}

\appendix

% \section{Appendix}
\label{sec:appendix}

\section{Task Setup}\label{sec:appendix task}
% \subsection{Task Setup}
LexNum consists of three tasks: economic compensation, work injury compensation, and traffic accident compensation.  

\textbf{Economic Compensation}: Based on the relevant provisions of the ``Labor Contract Law of the People's Republic of China'', this task involves determining eligibility for compensation, calculating the length of service and average monthly wage, and computing the economic compensation amount to ensure that employees receive lawful and reasonable compensation upon contract termination.  

\textbf{Work Injury Compensation}: This task identifies the responsible party and, in accordance with the ``Work Injury Insurance Regulations'' and other relevant laws, calculates various compensation amounts to ensure that injured employees receive lawful and reasonable compensation.  

\textbf{Traffic Accident Compensation}: This task determines liability, calculates reasonable damages based on relevant legal provisions, and determines the final compensation amount according to insurance coverage, ensuring that the injured party receives lawful and fair compensation.  

Examples of these three tasks are shown in Figure~\ref{fig:example}.
Economic compensation requires accurately determining the number of months for which compensation is due based on legal provisions, as well as the maximum amount that can be compensated.
Traffic accident compensation also requires distinguishing the contents included in different types of compensation. For example, the question in Figure~\ref{fig:example}(b) asks about compensation related to work-related death benefits, so funeral allowances should not be considered.
Work injury compensation requires identifying which costs in the problem are related to the specified compensation, while also understanding legal provisions. For example, the question in Figure~\ref{fig:example} (c) asks about compensation related to compulsory insurance, which has a limit. Considering medical expenses and allowances, the maximum compensation is 10,000.

% This is an appendix.

\section{Prompt Used in the Data Costruction}\label{sec:appendix prompt}
\begin{figure*}
    \centering
\includegraphics[width=0.98\linewidth]{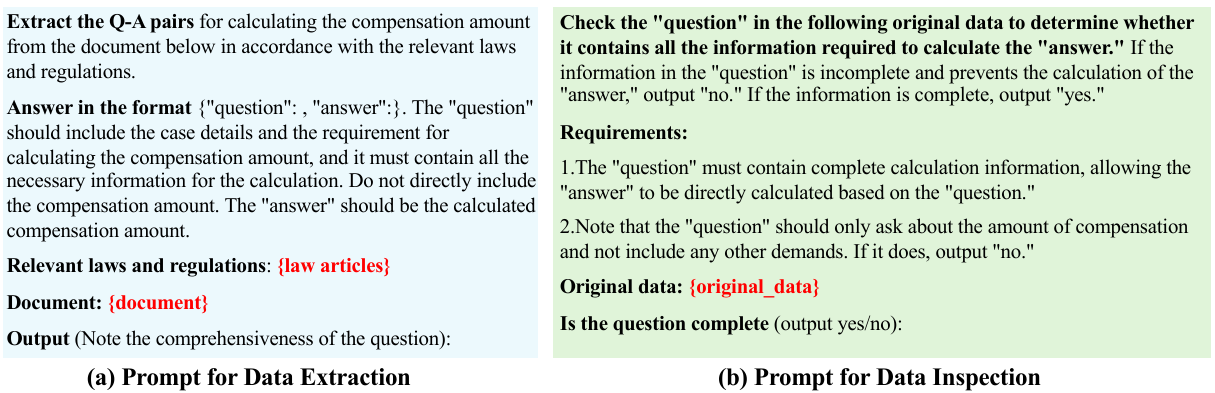}
    \caption{
    The prompts of the data costruction. The red parts are replaced according to the specific situation.
In (a), {law articles} is replaced with different "Relevant laws and regulations" depending on the dataset, and {document} refers to the specific legal document.
In (b), {original\_data} is the extracted JSON file, which contains the keywords "question" and "answer."
   }
    \label{fig:appendix prompt}
\end{figure*}
Figure~\ref{fig:appendix prompt} shows the specific prompt for us to construct the dataset using GPT-4o, which was written as suggested by the legal professional mentioned in §~\ref{sec:appendix annotator}.

\section{Specific Information of the Manual Annotator}\label{sec:appendix annotator}
We hired three legal annotators from the law program of a university in China. They are highly familiar with the legal provisions and calculation methods involved in the LexNum dataset provided in this paper. All three have undergone years of legal education and possess practical experience in the legal field. Among them, there are two females and one male, aged between 20 and 30. We explained to them that the reviewed dataset would be used for research on Chinese legal AI and provided reasonable compensation based on local standards.
We gave them detailed instructions on which legal provisions and relevant regulations should be used to review and revise the questions and answers. We also informed them that the dataset should be refined following the procedures described in §~\ref{sec:data quality}.

\section{More Details of the Evaluation LLM}\label{sec:appendix llm}
Table~\ref{tab:model urls and license} are the website URLs and corresponding licenses of the evaluated models.

\begin{table*}[t]
% \small
\centering
 % \vspace{-3px}
\resizebox{0.9\textwidth}{!}{
\begin{tabular}{cl|ll}
    \toprule
    Type&LLM &URL & Licence   \\

    \hline
    \multicolumn{1}{c}{\multirow {7}{*}{\shortstack{Legal Domain}}}
&fuzi.mingcha&\url{https://github.com/irlab-sdu/fuzi.mingcha} &Apache-2.0 license\\
    &DISC-LawLLM &\url{https://github.com/FudanDISC/DISC-LawLLM}&Apache-2.0 license \\
    &LawGPT\_zh&\url{https://github.com/LiuHC0428/LAW-GPT}& \\
    &Hanfei&\url{https://github.com/siat-nlp/HanFei}&Apache-2.0 license\\
    &Tailing&\url{https://github.com/DUTIR-LegalIntelligence/Tailing}&\\
    &LexiLaw&\url{https://github.com/CSHaitao/LexiLaw}&MIT license\\
    &zhihai&\url{https://github.com/zhihaiLLM/wisdomInterrogatory}&Apache-2.0 license\\
    \hline
    \multicolumn{1}{c}{\multirow {2}{*}{\shortstack{Open domain}}}&
QwQ-32B-Preview &\url{https://huggingface.co/Qwen/QwQ-32B-Preview}&Apache-2.0 license\\ 
    &DeepSeek-R1 &\url{https://github.com/deepseek-ai/DeepSeek-R1} &MIT license\\
    \bottomrule
\end{tabular}
}
\caption{
The LLM source URLs and licenses used by this paper. The parts where the license is listed as empty indicate that the author has not provided a License.}
\label{tab:model urls and license}
 % \vspace{-0.3cm}
\end{table*}

\section{Qualitative Analysis}\label{sec:appendix case study}
The four students who evaluated the model responses each had no less than three years of legal education in Chinese law and were familiar with the relevant statutes and legal procedures related to compensation calculation in Chinese legal contexts.
Among them, there were two females and two males, aged between 20 and 30. We explained to them that the evaluation was conducted to support research in Chinese legal AI and provided reasonable compensation based on local standards.
We gave them a detailed explanation of each evaluation criterion, including what specific aspects to consider, and used several examples to ensure that their evaluation standards were aligned.

\begin{figure*}
    \centering
\includegraphics[width=0.98\linewidth]{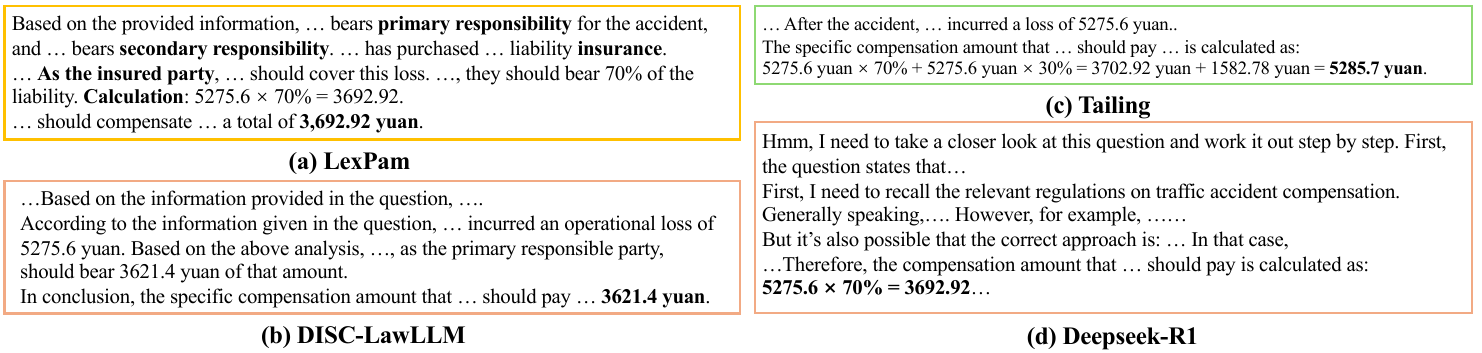}
    \caption{
    Responses from four models on the traffic compensation dataset.
   }
    
    \label{fig:appendix case}
\end{figure*}
Figure~\ref{fig:appendix case} presents example responses from four models on the traffic compensation dataset. We can observe that the legal LLM, having undergone SFT on legal QA tasks, is able to generate relatively concise responses to some extent. However, its calculation results are often incorrect, as the injection of extensive legal knowledge tends to impair its general mathematical reasoning ability.
DeepSeek-R1, on the other hand, performs well on mathematical computation tasks, but its reasoning process is often overly verbose, making it less suitable for user-friendly reading.
LexPam, by contrast, is able to produce relatively concise responses while also demonstrating strong mathematical computation capabilities.


% \input{latex/sections/appendix}

\end{document}